\pdfoutput=1

\documentclass[11pt]{article}

\usepackage{acl}
\usepackage{times}
\usepackage{latexsym}

\usepackage[T1]{fontenc}

\usepackage[utf8]{inputenc}
\usepackage[utf8]{inputenc} 
\usepackage[T1]{fontenc}    
\usepackage{booktabs}       
\usepackage{amsfonts}       
\usepackage{nicefrac}       
\usepackage{microtype}      
\usepackage{xcolor}         
\usepackage{pifont}
\usepackage{threeparttable}
\usepackage{multirow}
\usepackage{amsfonts}
\usepackage{bbm}
\usepackage{bm}
\usepackage{bbold}
\usepackage{color}
\usepackage{graphicx}
\usepackage{amsmath}
\usepackage{subfigure}
\usepackage{enumitem}
\newcommand\our{\makebox{\textsc{STEPS}}}
\usepackage{microtype}

\title{STEPS: A Benchmark for Order Reasoning in Sequential Tasks}

\author{Weizhi Wang, Hong Wang,  Xifeng Yan \\
  University of California, Santa Barbara, CA, United States \\
  \texttt{\{weizhiwang, hongwang600\}@ucsb.edu, xyan@cs.ucsb.edu}}

\begin{document}
\maketitle
\begin{abstract}
Various human activities can be abstracted into a sequence of actions in natural text, i.e. cooking, repairing, manufacturing, etc. Such action sequences heavily depend on the executing order, while disorder in action sequences leads to failure of further task execution by robots or AI agents. Therefore, to verify the order reasoning capability of current neural models in sequential tasks, we propose a challenging benchmark , named \our{}. \our{} involves two subtask settings, focusing on determining the rationality of given next step in recipes and selecting the reasonable step from the multi-choice question, respectively. We describe the data construction and task formulations, and benchmark most of significant Large Language Models (LLMs). The experimental results demonstrate 1) The commonsense reasoning of action orders in sequential tasks are challenging to resolve via zero-shot prompting or few-shot in-context learning for LLMs; 2) Prompting method still significantly lags behind tuning-based method on \our{}. The benchmarking dataset will be open-sourced at \url{https://github.com/Victorwz/STEPS}.
\end{abstract}

\section{Introduction}

Human tasks are universally described and abstracted into a sequence of actions. Such action sequences are mostly recorded and spread in the form of natural text, i.e. in the form of recipes, product manuals, service manuals, etc. Human have generalized and flexible capability to understand such action sequences and executing such actions in order. Human can also easily infer whether the given step is a reasonable next step without exposure to large amount of prior knowledge. With such reasoning capability, human can avoid making disordered next steps or actions to prevent the failure or accidents on the whole tasks. For example, boiling water should always go behind adding pasta to the pot, otherwise the pasta will get burned. Therefore, reasoning about the plausibility of next steps are essential for human to accomplish both daily and producing tasks.

\begin{figure}[tb] 
\centering 
\includegraphics[width=0.5\textwidth]{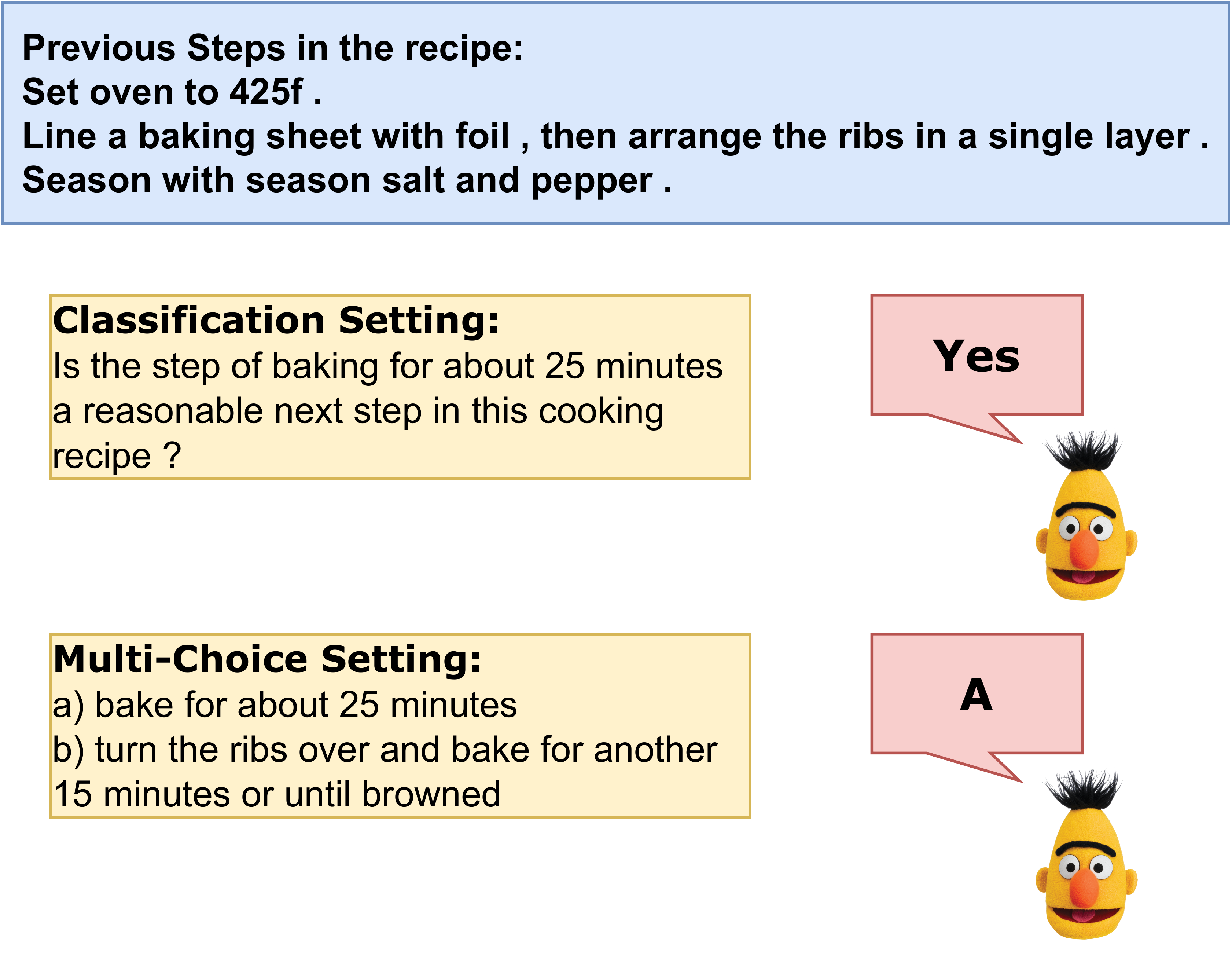} 
\caption{Illustration of two subtask settings of the proposed \our{} benchmark.} 
\label{fig:data} 
\end{figure}

Large Language Models (LLMs)~\cite{devlin2018bert,liu2019roberta,radford2019language,brown2020language} have significantly promoted the state-of-the-art on benchmarks of natural language understanding and generation~\cite{turingnlg,su2019generalizing,liang2022helm,wang2022task}. Enabled by self-supervised learning on large-scale high-quality training corpus and billions of parameters, LLMs are found to be capable of completing downstream tasks as few-shot or zero-shot learners. Via simple prompting with task-specific natural language templates, LLMs can achieve state-of-the-art performance on text classification, sentiment analysis, reading comprehension, language modeling, etc. without any further tuning with task-specific data. In addition, using the method in-context learning (ICL) to get LLMs exposed to several prompting task-specific examples, LLMs can harvest the task-specific knowledge in given local context and achieve human-parity performance on downstream tasks~\citep{brown2020language}.

Human can easily draw an answer to commonsense questions via access and memory to world knowledge and daily observations. However, LLM only encodes and acquires the commonsense knowledge via pre-training on large text corpora. Such knowledge is implicitly encoded in its trainable parameters, which weakens its reasoning capability without explicit memory and access towards commonsense knowledge. For example, human can easily order sequential actions in a recipe that pre-heating the oven goes before putting the pizza into it because human are heavily exposed to daily cooking scenarios and a large amount of knowledge bases like Internet or Books. In contrast, LLMs can only acquire such simple commonsense knowledge via neural-based memorization on extremely small split of recipes in web-crawled text dataset. To robustly and effectively evaluate the action order reasoning capability of language models, we propose a novel benchmark \our{}. \our{} involves two subtask settings: classification, which verifies the reasoning capability in determining the rationality of the  candidate next step given previous steps in a recipe, and multi-choice setting, which focuses on differentiating the correct next step choice given two candidate next steps and the previous steps in the recipe. Firstly, We formulate the sequence order reasoning evaluation into two evaluation tasks, the classification and multi-choice questions. Then we present the data resources, data construction and evaluation setting in details. Based on that, we benchmark three groups of state-of-the-art LLMs (GPT2, OPT, BLOOM) as baselines to evaluate their sequence order reasoning capabilities on recipes.

\section{Related Work}

\paragraph{Large Language Models for Reasoning Tasks.}
LLMs are becoming dominant methods on all natural language processing tasks in the few-shot or zero-shot learning manner, while their capabilities in commonsense reasoning remain under explored. \citet{qiao2022reasoning} classifies the various reasoning benchmarks into four categories based on the required reasoning skills, arithmetic reasoning~\citep{qian2022limitations,mishra2022lila}, commonsense reasoning~\cite{talmor2018commonsenseqa,bisk2020piqa}, logical reasoning~\cite{dalvi2021explaining}, symbolic reasoning~\cite{wei2022chain}, and multimodal reasoning~\citep{wang2022visually}. The proposed Sequence Order Reasoning benchmark lies at the research line of commonsense reasoning for LLMs. In addition to conventional directly answering methods of fully fine-tuning, zero-shot prompting or in-context learning, Chain-of-Thought (CoT) Prompting~\cite{wei2022chain} guides LLMs to generate explicit intermediate reasoning steps to get the final answer.


\section{\our{}}
\subsection{Task Formulation}
\paragraph{Classification Setting.} The task of next step reasoning requires the LLMs to figure out whether the given step is a reasonable next step and is confronted with the previous recipe. The task can be naturally formulated into a binary classification task on the given textual concatenation of previous steps and candidate next step, which can be formulated as follows:
Assume a recipe $R_i\in \mathcal{D}_R$ contains $N_i$ action steps described in textual sentences $\{S^{(i)}_{1}, S^{(i)}_{2}, \cdots, S^{(i)}_{N_i}\}$, $\forall j\in [2,\cdots,N_i-1]$. Given the previous steps $\{S^{(i)}_{1}, S^{(i)}_{2}, \cdots, S^{(i)}_{j-1}\}$, each candidate step in the set of $\{S^{(i)}_{j}, S^{(i)}_{j+1}, \cdots, S^{(i)}_{N_i}\}$ will be classified and recognized based on whether it is the correct next step given the previous steps. The ground truth next step $S^{(i)}_j$ should be classified into the label of "Yes" while the other candidates $S^{(i)}_{j+1}, \cdots, S^{(i)}_{N_i}$ should be classified into "No". 

\paragraph{Multi-Choice Question Setting.} Parallel to the classification task setting which lies at the natural language understanding pattern, we propose the second task setting, multi-choice question setting, which is more confronted with the causal language modeling manner of LLMs. Given the textual sequence of the previous steps $\{S^{(i)}_{1}, S^{(i)}_{2}, \cdots, S^{(i)}_{j-1}\}$, LLMs are required to choose the correct next reasonable step in the two step candidates. The correct choice of next step will be the step of $S^{(i)}_j$, while the false choice is a step randomly selected from the steps $\{S^{(i)}_{j+1}, \cdots, S^{(i)}_{N_i}\}$.

\subsection{Dataset Construction}
The benchmark of sequence order reasoning is conducted based on the \textsc{Food.com Recipes} dataset, a web crawled dataset of recipes from ``food.com`` over 18 years which is collected and released by~\citet{majumder2019generating}. We keep the original train/dev/test splits for recipes and filter the recipes with less than four action steps or more than ten action steps to avoid exceeding input context length limitation of LLMs. To construct the classification dataset, for each original recipe with $N$ steps, the true sample are constructed as $(S_{[1:i-1]}, S_i),\forall i\in  [2,N-1]$ and the false samples are constructed as $(S_{[1:i-1]}, S_i),\forall i\in  [2,N-1], \forall k\in [j+1, N] $. To construct the dataset of multi-choice question setting, for each original recipe with $N$ steps, each sample is constructed as the tuple of $(S_{[1:i-1]}, S_i, S_k),\forall i\in  [2,N-1]$, in which $k$ is a random selection within $[i+1, N]$.
The statistics for the constructed datasets on two subtask settings are presented in Table~\ref{tab:stat}.

\begin{table}
\centering
\begin{tabular}{l | ccc}
\hline
\toprule
\textbf{Data} & \textbf{Train} & \textbf{Dev} & \textbf{Test}\\
\midrule
\our{}-CLS & 2.68M & 1K & 1K \\
\our{}-MC & 298K & 1K & 1K \\
\bottomrule
\hline
\end{tabular}
\caption{Dataset statistics for the two proposed task settings in \our{} benchmark.}
\label{tab:stat}
\end{table}

\subsection{Baselines}
For the baselines of proposed \our{} benchmark, we evaluate three groups of significant large language models, including 1) four size of GPT-2~\cite{radford2019language} (Small, Medium, Large, and XL); 2) three size of Open-Pretrained-Transformer Language Models (OPT)~\cite{zhang2022opt} (1.3B, 13B, and 30B); 3) two size of BigScience Large Open-science Open-access Multilingual Language Model (BLOOM)~\cite{scao2022bloom} (3B, 7B1). We include the model architecture details for all baseline LLMs in Table~\ref{tab:baseline}.

\begin{table}[ht]
\small
\begin{tabular}{lccccc}
\hline
\toprule
\textbf{Models} & \textbf{\#Params} & \textbf{\#L} & \textbf{Embd} & \textbf{\#H} & \textbf{Bsz}\\
\midrule
GPT2-Small & 117M  & 12 & 768 &  12  & 0.5M \\
GPT2-Medium & 345M & 24 & 1024  & 16  & 0.5M \\
GPT2-Large & 774M  & 36  & 1280 & 20  & 0.5M \\
GPT2-XL  & 1.5B & 48 & 1600 & 25 & 0.5M \\
OPT-1.3B & 1.3B & 24  & 2048 &  32 & 1M \\
OPT-13B & 13B & 40 & 5120 & 40  & 1M \\
OPT-30B & 30B & 48 & 7168 & 56 & 4M \\
BLOOM-3B & 3B & 30 & 2560 & 32 & 1M \\
BLOOM-7B1 & 7.1B & 30 & 4096 & 32 & 1M \\
\bottomrule
\hline
\end{tabular}
\caption{Model architecture details of baseline LLMs evaluated on the proposed benchmark. We present the number of parameters (\#Params), the number of layers (\#L), Embedding Dimension (Embd), the number of heads (\#H), and the training batch size in tokens (Bsz).}
\label{tab:baseline}
\end{table}

\begin{table}
\centering
\begin{tabular}{l |  c  }
\hline
\toprule
\textbf{Model (\#Params$\uparrow$)} & \textbf{Multi-Choice Accuracy$\uparrow$} \\
\midrule
Majority Class & 50 \\
\midrule
GPT2-Small  & 59.3 \\
GPT2-Medium & 63.5 \\
GPT2-Large  & 65.1  \\
OPT-1.3B & 66.0 \\
GPT2-XL & 65.4 \\
BLOOM-3B & 68.4 \\ 
BLOOM-7b1 & 69.8 \\
OPT-13B   & 71.3 \\
OPT-30B &  71.7 \\
\bottomrule
\hline
\end{tabular}
\caption{
The results of baseline large language models on the proposed \our{} benchmark. We report accuracy as the evaluation metric. For the results on few-shot in-context learning, we adopt six random seeds to select the demonstration examples. We report the mean of the accuracy on six random seeds and attach the standard deviation as the subscript. 
}
\label{tab:qa}
\end{table}

\begin{table*}[!htb]
\centering
\small
\begin{tabular}{l | ccc | ccc|ccc  }
\hline
\toprule
\multirow{2}{*}{\textbf{Model}} & \multicolumn{9}{c}{\textbf{Classification Task Performance [\%]}}  \\
& \multicolumn{3}{c}{\textbf{Zero-shot Inference}} & \multicolumn{3}{c}{\textbf{Few-Shot ICL}} & \multicolumn{3}{c}{\textbf{Fine-Tuning}} \\
\midrule
\# Demons. & \multicolumn{3}{c|}{0} & \multicolumn{3}{c|}{4} & \multicolumn{3}{c}{0} \\
\midrule
Metrics & Sen. & Sepc. & G-Mean & Sen. & Sepc. & G-Mean & Sen. & Sepc. & G-Mean \\
\midrule
Random & 48.6 & 51.2 & 49.9 & 48.6 & 51.2 & 49.9 & 48.6 & 51.2 & 49.9 \\
\midrule
GPT2-S  & 95.4 & 5.3 & 22.6 &   $98.5_{2.2}$  & $1.8_{2.4}$ & $8.7_{9.9}$  & 81.2 & 72.8  & 76.9  \\
GPT2-M & 100.0 & 0.4 & 6.1 & $90.1_{14.8}$ & $10.1_{12.0}$ & $22.2_{15.8}$ & 90.8 & 68.3 & 78.8 \\
GPT2-L  & 99.5 & 0.4 & 6.1 & $76.4_{11.7}$ & $29.6_{13.5}$ & $44.5_{11.4}$  & 83.0 & 73.4 & 78.1 \\
GPT2-XL     & 100.0 & 0.4  & 6.1 & $78.5_{11.4}$ & $25.6_{11.7}$ & $42.5_{8.8}$ & -  & - & -\\
\midrule
OPT-1.3B    & 49.1 & 60.5 & 54.5 &  $62.8_{9.7}$  & $39.2_{11.2}$ & $48.4_{4.8}$   & -  & - & -  \\ 
OPT-13B     & 84.9 & 20.4 & 41.6 &  $44.0_{15.0}$  & $68.0_{10.6}$ & $52.9_{7.3}$   & -  & - & -  \\
OPT-30B     & 85.8  & 14.5  & 35.3  &  $86.7_{14.4}$  & $16.7_{14.4}$ & $33.6_{10.8}$  & -  & - & -  \\
\midrule
BLOOM-3B    & 99.5 & 1.2 & 11.1 & $97.3_{1.9}$ & $5.8_{3.8}$ & $22.5_{7.4}$   & -  & - & - \\
BLOOM-7B1   & 75.7 & 32.7 & 49.7 & $98.4_{3.6}$  & $1.3_{2.3}$ & $8.5_{7.2}$  & -  & - & - \\
\bottomrule
\hline
\end{tabular}
\caption{
Experimental results of baseline LLMs on classification subtask of the proposed \our{} benchmark. We report the classification Sen. (Sensitivity), Spec. (Specificity), and G-Mean (Geometric Mean of Sensitivity and Specificity) of each baseline LLM, in which Sensitivity and Specificity are the class-wise accuracy on positive and negative sets respectively. For the results on few-shot in-context learning, we adopt six random seeds to select the demonstration examples. We report the mean of the metric on six random seeds and attach the standard deviation as the subscript. 
}
\label{tab:res}
\end{table*}

\subsection{Evaluation Setting}
\paragraph{Classification Evaluation Setting.} All baseline large language models are evaluated in zero-shot, one-shot and few-shot in-context learning manner. For the few-shot in-context learning evaluation, we set the number of demonstration examples $K$ in the prompt context as $4$. The $k$ demonstration examples are balanced, in which $k/2$ positive samples and $k/2$ negative samples are randomly selected from positive and negative training samples, respectively. For each $K$ we evaluate with six random seeds for the random selection on the demonstrations and the mean and standard deviation of classification accuracy are reported. For each test sample with (previous-steps, next-step, label), we deploy the textual task template \textit{"[previous-steps] . Is the next step of [next-step] a reasonable step in this recipe ?"} to concatenate the previous steps and the candidate step into a prompt query. Then the prediction label is chosen based on the relation between $P(\text{Yes}|\text{query})$ and $P(\text{No}|\text{query})$.
In addition, we fully fine-tune three size of GPT-2 models on the training set of the classification subtask to verify the task-specific adaptation capability of LLMs, and the fine-tuning details are presented in Appendix~\ref{sec:ft}. As the constructed dataset for classification task is imbalanced (78.7\% negative versus 21.3\% positive samples), we deploy the class-wise accuracy as the evaluation metrics for classification task setting. Specifically, the classification accuracy on positive class (\texttt{Sensitivity}), the classification accuracy on negative class (\texttt{Specificity}), and the geometric mean of \texttt{Sensitivity} and \texttt{Specificity} are deployed towards the proposed imbalanced classification task. To avoid imbalanced fine-tuning, we adopt up-sampling on the positive samples in the training set to match the number of negative samples. For each fine-tuned baseline LLM, we truncate the whole resampled training set into segments, of which the length is 1024 tokens. We fine-tune each LLM for 6000 updates in total with the batch size of 8. We perform the validation every 300 updates and save the checkpoint with best performance on validation set. The LLM checkpoints are accessed via \texttt{Huggingface transformers}~\cite{wolf2019huggingface}. We deploy Adam \citep{kingma2014adam} ($\beta_1=0.9,\beta_2=0.98$) optimizer and train all models with $lr=0.0003$.


\paragraph{Multi-Choice Question Evaluation Setting.} For each test samples with the previous steps and two options, each query $q$ into LLM is the concatenation of the previous steps and one option. We follow~\citet{radford2019language} to use the language modeling perplexity as the scorer for each potential query  $\textnormal{score}(q)=\text{PPL}(\text{q})$. Then the option with lower language modeling score is selected as the solution. The multi-choice answering accuracy is used as the evaluation metric in this task setting.

\subsection{Benchmark Results}
The evaluation results of baseline LLMs on the proposed two subtasks of \our{} benchmark are presented in Table~\ref{tab:qa} and Table~\ref{tab:res}. 
\paragraph{Classification Results.} In zero-shot prompting evaluation, we find that GPT-2 series LLMs partially fail in performing the classification on the rationality of given next steps. The model predictions all lies at positive class towards 1000 testing samples, leading to perfect sensitivity but almost 0 specificity. In addition, it is hard to conclude that the performance strictly increases with the scaling up of the model. Scaling up the model is beneficial to performance improvement for BLOOM models, in which BLOOM-7b1 significantly outperforms its 3B size model by 38.6\% G-Mean score. But such scaling law is not supported for GPT-2 and OPT series of models and the largest baseline LLMs, OPT-30B performs worse than smaller OPT models, OPT-13B. Secondly, providing demonstrations to perform in-context learning helps LLMs to avoid fully biased predictions on positive class, in which three size of GPT-2 models (M, L, XL) gain large performance improvement on both \textsc{Specificity} and G-Mean score. However, the demonstration examples do not contribute to the LLMs which have performed well in zero-shot learning, including OPT-1.3B, OPT-30B, and BLOOM-7B1. At last, tuning-based method still achieves best performance compared with non-parametric methods for each LLM. The balanced fine-tuning can effectively fix the issues of biased prediction and zero \textsc{Specificity} for GPT-2 (S, M, XL) models, leading to the increase of 73\% \textsc{Specificity} for GPT2-Large compared with zero-shot learning.

\paragraph{Multi-Choice Results.} The results on multi-choice answering subtask for LLMs strictly follow the scaling law, in which the largest model, OPT-30B outperforms all others with 71.7 answering accuracy. In addition, we find that such scaling law is even valid across different groups of LLMs, in which BLOOM-3B achieves better performance than OPT-1.3B. Such experimental results demonstrate that the multi-choice question is a more effective, accurate, and robust method for evaluating LLMs because it is evaluated via language modeling perplexity scoring which is the same as pre-training objective of LLMs.

\section{Conclusions and Discussions}
In this paper, we propose a novel commonsense reasoning benchmark \our{} for order reasoning in sequential tasks. We present the two evaluation subtask settings for \our{}, classification task and multi-choice question answering task, as well as the task formulation and data construction. We benchmark most of state-of-the-art LLMs on \our{} for further comparisons.

Overall, the experimental results demonstrate that the performance of LLMs on both classification and multi-choice question settings all lie at the interval of 70\%-80\% using the accuracy-style metrics, which might be a potential performance upper-bound for LLMs pre-trained on large-scale corpora via self-supervision. To go beyond this performance bound, more commonsense knowledge bases and effective chain-of-thought prompting method are supposed to be introduced in LLMs reasoning on the proposed sequence order reasoning benchmark.

\section{Acknowledgements}
This research was partly sponsored by the DARPA PTG program (HR001122C0009). Any opinions, findings, conclusions, or recommendations expressed in this paper are those of the authors and do not necessarily reflect the views of funding agencies.


\bibliography{custom}
\bibliographystyle{acl_natbib}





\end{document}